\documentclass{article}
\usepackage[utf8]{inputenc}
\usepackage{amsmath}
\usepackage{amsthm}
\usepackage{amssymb}
\usepackage{amsfonts}
\usepackage{graphicx}
\usepackage{optidef}
\usepackage[margin=1in]{geometry}
\usepackage{float}
\usepackage{authblk}
\usepackage[ruled,vlined,linesnumbered]{algorithm2e}
\usepackage{hyperref}
\usepackage{mathrsfs}

\usepackage[T1]{fontenc}
\usepackage{multirow}
\usepackage{natbib}
\title{Federated Automated Feature Engineering}
\author[2]{Tom Overman$^1$, Diego Klabjan$^2$}
\date{$^1$Department of Engineering Sciences and Applied Mathematics, Northwestern University\\%
      $^2$Department of Industrial Engineering and Management Sciences, Northwestern University\\[2ex]
}

\begin{document}

\maketitle

\section{Abstract}
Automated feature engineering (AutoFE) is used to automatically create new features from original features to improve predictive performance without needing significant human intervention and domain expertise. Many algorithms exist for AutoFE, but very few approaches exist for the federated learning (FL) setting where data is gathered across many clients and is not shared between clients or a central server. We introduce AutoFE algorithms for the horizontal, vertical, and hybrid FL settings, which differ in how the data is gathered across clients. To the best of our knowledge, we are the first to develop AutoFE algorithms for the horizontal and hybrid FL cases, and we show that the downstream test scores of our federated AutoFE algorithms is close in performance to the case where data is held centrally and AutoFE is performed centrally.

\section{Introduction}
Automated feature engineering (AutoFE) aims at automatically creating and selecting new features from the original set of features that improve downstream predictive performance. Feature engineering is often used to craft new features that help improve simple models. These crafted features may include nonlinear transformations of existing features that allow linear models to capture more complex, nonlinear interactions. Traditionally, feature engineering required specialized domain knowledge and extensive time for trial-and-error testing. However, recent approaches in automated feature engineering allow data scientists to automatically craft new informative features with very limited time, effort, and domain knowledge.

Federated learning (FL) is a privacy-preserving machine learning technique that has gained traction in recent years. The main requirement in federated learning is that data belonging to client devices \textit{cannot} be directly sent to the server to build a global model. Instead, the typical approach is for each client to build a local model with only their own local data, then the clients can send their model weights to the server which aggregates these weights in various ways. Many of the key challenges in federated learning revolve around minimizing the number of communications between the clients and server and maintaining good performance when the data across clients are non-IID \cite{fl}.

There are three federated learning settings that correspond to how the data is gathered across clients. In horizontal federated learning, each client has a subset of samples and each of these samples has all of the features \cite{fl}. In vertical federated learning, each client has the full set of samples but only a subset of the features \cite{fedbcd}. Finally, in hybrid federated learning, each client only has a subset of samples and each of these samples only has a subset of features \cite{hyfdca}\cite{hyfem}\cite{hybrid-graph}. Each of these settings requires different algorithmic approaches.

This paper focuses on automated feature engineering in the context of federated learning, where clients are not able to send data to the server, but we still build informative new features from the original raw features. We introduce algorithms for all three FL settings: horizontal, vertical, and hybrid. The horizontal algorithm we propose extends IIFE \cite{iife} to the federated setting, which we term Fed-IIFE. The hybrid and vertical FL methods hinge on homomorphic encryption and differential privacy to safely form combinations of features that reside on different clients, along with a feature selection scheme to determine which features to keep. All of the algorithms are iterative in that they gradually construct more and more complicated engineered features each iteration.

Our contributions are as follows
\begin{enumerate}
    \item To the best of our knowledge, we are the first to develop federated AutoFE algorithms in the horizontal and hybrid FL settings. We also provide a new vertical AutoFE algorithm that is safer from client privacy leaks than the one existing approach in this setting.
    \item We provide experiments demonstrating that our horizontal FL AutoFE algorithm performs similarly to the case where AutoFE is performed centrally, even in cases where the data distributions across clients are highly non-IID. We validate the ability of federated interaction information to determine synergy between feature pairs with a synthetic experiment. We also demonstrate that the important engineered features generated by centralized IIFE have similarly generated features in Fed-IIFE, even in highly non-IID settings.
    \item We demonstrate that our hybrid and vertical AutoFE algorithms still show significant improvements over the baseline (no engineered features) even with the strictly limiting requirements of homomorphic encryption and differential privacy.
\end{enumerate}

In Section \ref{sec:related_work}, we review the relevant work on both federated learning and AutoFE and discuss the minimal and incomplete work that exists on AutoFE in the vertical FL setting. In Section \ref{sec:algorithms}, we detail our proposed AutoFE algorithms for the horizontal, vertical, and hybrid FL settings. In Section \ref{sec:experiments}, we show experimental results for all three FL settings, and verify key aspects of Fed-IIFE such as the validity of federated interaction information and similarity of engineered features between the centralized and FL settings.

\section{Related Work}
\label{sec:related_work}
There is significant work existing on automated feature engineering. The main focus of most algorithms is on cleverly navigating the massive search space of possible combinations of features for high order engineered features. OpenFE uses a very fast boosting technique based on gradient boosted trees to quickly evaluate all possible feature combinations \cite{openfe}. AutoFeat enumerates all possible combinations up to a specified order, then uses a fast feature selection technique to select the best performing features \cite{autofeat}. IIFE tackles this problem by using interaction information to determine the best synergizing feature pairs and only exploring the best pairs of features, then iteratively builds higher order features \cite{iife}. EAAFE uses a genetic algorithm approach to search the best engineered features \cite{eaafe}. DIFER uses a deep learning approach to find representations of the engineered feature strings and their predicted validation scores \cite{difer}. Furthermore, very recently there have been a focus on using Large Language Models to build engineered features in an informed fashion based on feature descriptions \cite{caafe}.

Federated learning also has extensive algorithmic work in addition to interesting theoretical results. Federated averaging is the original FL algorithm that is designed for the horizontal setting \cite{fedavg}. Further improvements for heterogeneous data exist, such as SCAFFOLD \cite{scaffold} and RadFed \cite{radfed}. The vertical FL setting has also had significant work in recent years \cite{fedbcd}. The hybrid federated learning case where clients have subsets of both samples and features has far fewer algorithms, but has recently had more work exploring this case \cite{hyfdca}\cite{hyfem}\cite{hybrid-graph}.

Despite the great deal of work in federated learning and automated feature engineering, there is relatively limited work on automated feature engineering in the federated learning setting. All of the existing work at the time of this writing focuses on the vertical case, where each client has different features, and all clients have the same samples. There exists work in AutoML in a federated setting, however automated feature engineering is not one of the aspects discussed in this work \cite{autofl}. FLFE is an algorithm for vertical federated AutoFE that uses quantile sketching to encode features and uses deep neural networks to predict if a particular transformation of features is useful \cite{vertfefl}. The neural networks are trained on a number of datasets, but these datasets are not shared, so it is impossible to reproduce their results. Since these datasets are actually used to build the feature engineering framework and not just used for evaluating the algorithm, it is necessary to have these datasets to properly evaluate the algorithm. Finally, homomorphic encryption is used when constructing the best candidate features for building the final model. However, this work shows only small gains compared to using only the raw features, does not compare to AutoFE methods in the centralized setting, and does not provide any source code making reproducibility very difficult. Furthermore, clients may be able to reconstruct the features of other clients from the decrypted engineered features; we address this in our work by utilizing differential privacy. We believe this makes the work very difficult to reproduce and compare. It is clear that there is a large research gap in all three settings of federated AutoFE: horizontal, vertical, and hybrid. Our work provides algorithms for all of these settings.

\section{Algorithms}
\label{sec:algorithms}
Adapting existing AutoFE algorithms to the horizontal FL setting is not straightforward. Many of these algorithms rely on expanding a very large number of engineered features and evaluating each of these new features, such as OpenFE, EAAFE, and AutoFeat. These extreme number of FL evaluations result in too much communication overhead to be practical. Furthermore, many algorithms have operations that are not straightforward to extend to the FL case, such as the LightGBM boosting evaluations that OpenFE performs. 

An alternative to directly incorporating FL into the algorithm is to perform AutoFE on each client separately, send the feature transformation strings to the central server where it takes the union of these transformations, and then send this union back to the clients. An optional step of performing federated feature selection after finding this union can further help performance; we provide such a federated feature selection algorithm in Section \ref{app:alternative_algos} of the Appendix. However, we find that this approach does not perform well because the feature engineering step does not incorporate any information from the other clients. This performance degradation is especially apparent when data is distributed across clients in a non-IID manner. We discuss this further in Section \ref{sec:fullyindependent} of the Appendix.

Fortunately, extending IIFE to the FL setting is relatively straightforward and results in good empirical performance compared to centralized IIFE, which is one of the top-performing AutoFE algorithms at the time of this writing \citep{iife}. We detail this algorithm, Fed-IIFE, in Section \ref{sec:horizontal_algo}. The cases of vertical and hybrid FL are more complicated than horizontal FL and are discussed in Section \ref{sec:vertical_algo}.

\subsection{Horizontal Setting}
\label{sec:horizontal_algo}
We extend IIFE to the FL setting by replacing some key aspects of IIFE with federated versions. We call this new FL version Fed-IIFE. Refer to the full algorithmic description of IIFE in \cite{iife}. We replace computations of interaction information with federated interaction information described in Algorithm \ref{Fed-II}. The key change is that interaction information is computed separately on each client. The clients send these interaction information scores to the server; the server then averages the contributions from all of the clients. The averaging can either be the simple average or a weighted average where the weights are the proportion of samples belonging to the client. We use the weighted average in our experiments. The other key change from IIFE to Fed-IIFE is that all model evaluations are performed in a federated fashion. Candidate transformations are computed on each client separately and then evaluated across all clients using any horizontal FL algorithm. In our experiments, we use FedAvg \citep{fedavg} for horizontal FL training.

The motivation for why we choose to extend IIFE to a federated version is that interaction information is straightforward to extend to the horizontal FL setting as shown in Algorithm \ref{Fed-II} and performs well even in highly non-IID settings as shown in Section \ref{results:fed-ii-validation}. Furthermore, the number of model training and evaluations is significantly reduced because interaction information significantly reduces the transformation space to explore. This cuts down on the number of communications needed during federated AutoFE while still yielding informative engineered features as shown in Section \ref{results:horizontal}.

\begin{algorithm}[htbp]
\label{Fed-II}
\SetAlgoLined
Notation: For the set $\mathcal{I}$ of tuples $((\tau_{ij},i,j))$, $\mathcal{I}[i,j]$ returns $\tau_{ij}$ of the tuple corresponding to the indices $i,j$.\\
Input: For all clients $c=1,..,Q$ sets $\mathcal{P}_c$ of feature pairs $(F^c_i, F^c_j)$ where $F^c_i$ indicates the feature $i$ with samples belonging to client $c$, Set $\mathcal{D}$ of corresponding index pairs $(i,j)$, and labels/targets $Y^c$ belonging to client $c$.\\
\textbf{Clients Execute:}\\
\For{clients $c=1,...,Q$}{
    $\mathcal{I}_c \gets \emptyset$ \\
    \For{$(F_i^c, F_j^c) \in \mathcal{P}_c$}{
        $\tau_{ij}^c=I(F_i^c,F_j^c,Y^c) = I(F_i^c,F_j^c|Y^c) - I(F_i^c, F_j^c)$ \\
        $\mathcal{I}_c \gets \mathcal{I}_c \cup \{(\tau_{ij}^c,i,j)\}$
    }
    Send $\mathcal{I}_c$ to server
}
\textbf{Server Executes:}\\
$\mathcal{I} \gets \emptyset$ \\
\For{$(i,j)$ in $\mathcal{D}$}{
    $\tau_{ij}=\underset{ c \in {1,...,Q}}{mean} \mathcal{I}_c[i,j]$ \tcp{Mean over all clients}
    $\mathcal{I} \gets \mathcal{I} \cup \{(\tau_{ij},i,j)\}$
}
return $\mathcal{I}$
\caption{Federated Interaction Information (Fed-II)}
\end{algorithm}

\subsection{Vertical and Hybrid Settings}
\label{sec:vertical_algo}
Vertical AutoFE algorithms must significantly differ from centralized and horizontal FL settings due to the fact that each client holds different feature spaces. The key aspect of AutoFE is combining different features together in ways that improve downstream model performance. However, directly sharing features on different clients with each other or a central server directly violates the requirements of FL. Furthermore, many vertical and hybrid FL algorithms require homomorphic encryption to protect privacy leakages when sharing intermediate information (such as inner products) for computing gradients \citep{hyfdca}. This is computationally expensive which severely limits the number of model training/evaluations that can be performed (AutoFE algorithms typically require a very large number of such trainings/evaluations). In fact, we show in Section \ref{sec:vertical_results} that homomorphic encryption is the main computational bottleneck in the vertical and hybrid algorithms.

\begin{figure*}[htbp]
\centerline{\includegraphics[width=.99\textwidth]{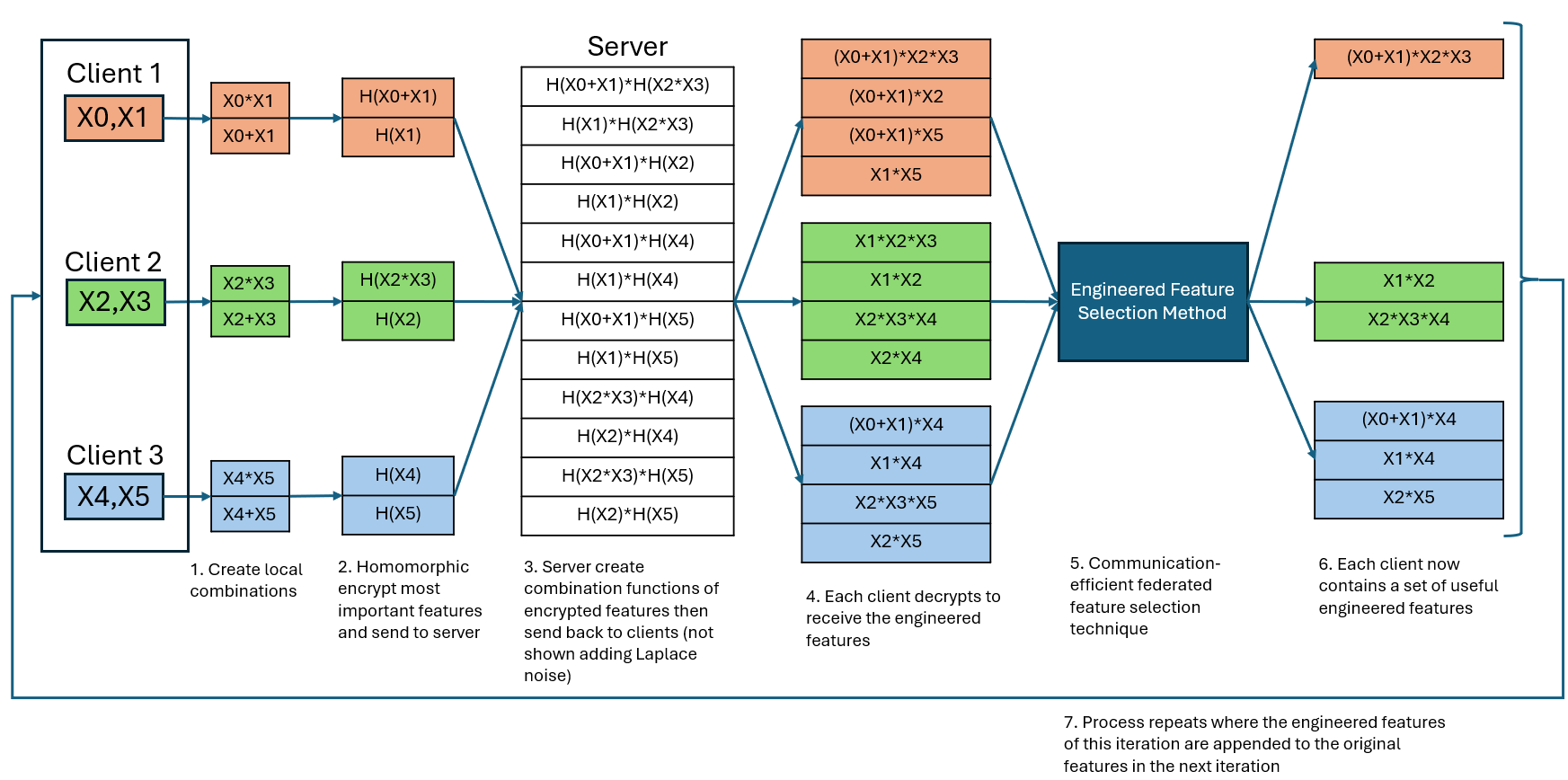}}
\caption[Workflow of Vertical-FLAFE]{Workflow of Vertical-FLAFE with 3 clients and each have two starting features. Typically there are more clients and the allowed function transformations are more complex.}
\label{vertical_flafe_diagram}
\end{figure*}

With these limitations in mind, we propose Vertical-FLAFE as shown in Algorithm \ref{vertical-flafe} and depicted in Figure \ref{vertical_flafe_diagram}. The basic idea for Vertical-FLAFE is that for features that are on separate clients and can't be directly shared, homomorphically encrypt the vectors, combine the features in the encrypted space, add encrypted Laplace noise (the server can do this with a public key, but is not able to decrypt), and finally send the combined features back to the clients to decrypt. This combined feature can then be evaluated to see if it is a useful new feature. Full details of privacy considerations are provided in the Appendix.

The homormorphic encryption schemes that are fast enough for practical use only preserve addition and multiplication operations. Thus, the set of candidate functions that exist in this setting is much smaller than the centralized or horizontal FL settings. However, since homomorphic encryption is very costly, especially for large datasets, we want to minimize the number of times we need this operation. This is why we first find the feature importance of each feature held on the client, and only the most important features on this client are combined with the most important features on other clients. 

Once all of the engineered features are created, sent back to clients, and decrypted, the best engineered features are selected using any feature selection technique that abides by FL requirements. If significant computational resources are available, then a federated feature selection approach, such as the one we propose in Section \ref{app:alternative_algos} of the Appendix, can be used to very accurately select the best features. However, in the typical case of constrained resources, then a fully local feature selection that determines the best engineered features on each client separately can be used. We use this approach in our experiments and provide details in Section \ref{vertical_results}.

It is important to note that most homomorphic encryption schemes require specifying the downstream function while encrypting the inputs, while in our approach we apply various functions (combinations of multiplications and additions) to the encrypted inputs. Fortunately, BFV and CKKS homomorphic encryption do not require the specification of the downstream function while encrypting and fit well within our approach \citep{bfv1,bfv2,ckks}.

\begin{figure*}[htbp]
\centerline{\includegraphics[width=.65\textwidth]{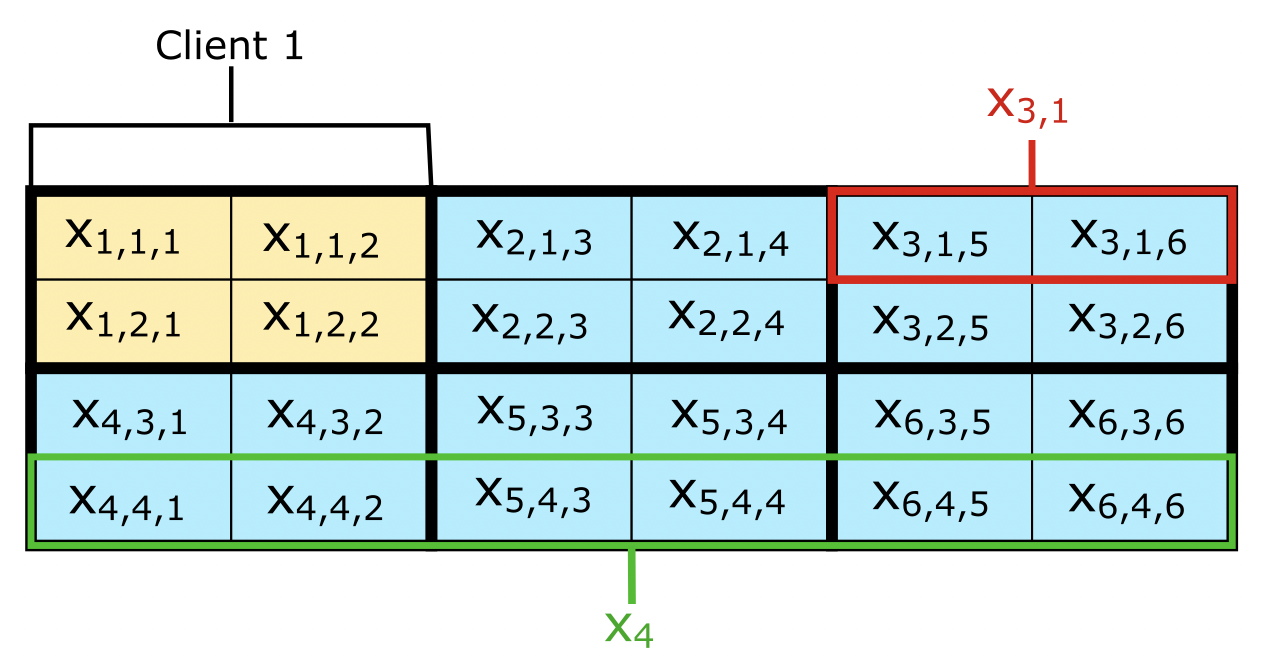}}
\caption[Example of how data is assumed to be partitioned for Hybrid-FLAFE]{Example of how data is assumed to be partitioned for Hybrid-FLAFE.}
\label{client_diagram}
\end{figure*}

We make some assumptions on the way that data is gathered across clients in the hybrid setting. While hybrid training algorithms, such as that by \cite{hyfdca}, allow fairly flexible partitions of data as long as each client hold rectangular patches of the overall data, we enforce more strict rules on the way data is split. Specifically, we require that every sample be partitioned across clients in the same way. Essentially, the vertical divisions must run across the entire sample space without any jagged partitions. An example of a suitable partition is shown in Figure \ref{client_diagram}. 

Hybrid-FLAFE follows very similarly to Vertical-FLAFE, but has some key differences that are needed in the hybrid setting. In Algorithms \ref{vertical-flafe}, \ref{createlocal}, and \ref{createnonlocal}, we denote which steps differ for the hybrid and vertical cases. First, the process of finding the most important features on each client to combine with other client's features needs to be done across several clients because each feature is held across many clients in the hybrid setting. This means feature importances must be computed across all clients that hold the same feature. This can be done through horizontal FL mechanisms or by simply averaging feature importances across all clients that share the same feature. We use the latter option in our experiments. Furthermore, any model training requires hybrid FL training algorithms, of which there are very few and these typically have assumptions on the type of models they work for. For example, as far as we know, there are no hybrid FL algorithms for tree-based models.

Algorithm \ref{createnonlocal} demonstrates how feature combinations from different feature spaces are performed. In the vertical case, after the combination is formed, it is straightforward to randomly select one of the two clients that owned the original features to send the new feature. However, in the hybrid case, the sample space is also partitioned, so it is possible that a combined feature $H^1_{i,j}$ is sent to the client holding original feature $i$ and later (in the different sample space) a combined feature $H^2_{i,j}$ is sent to the client holding original feature $j$. This would result in the overall data matrix becoming jagged and would cause errors in the subsequent feature selection and hybrid FL training. We handle this by keeping track of which group of clients that contain the same features we send this new feature to, and for subsequent combinations we assign the client to this same group to keep the feature space properly aligned.

\begin{algorithm}[htbp]
\caption{Vertical-FLAFE (Hybrid-FLAFE)}
\label{vertical-flafe}
\textbf{Input}: Clients $k=1,2,...,Q$, Local data matrix $X_k\in \mathbb{R}^{N_k\times M_k}$ and targets $y_k\in \mathbb{R}^{N_k}$ belonging to client $k$, CKKS Homomorphic encryption function $H(\cdot)$\\
\For{each outer iteration $i=1,...,O$}{
    \For{each client $k=1,...,Q$}{
       $\mathcal{F}_k = [ \;]$\\
       $\mathcal{F}_k = \mathcal{F}_k \cup \text{CreateLocalCombinations}(X_k)$ \tcp{Create combinations of features located on client}
    }
       \textbf{Vertical}:\\ \Indp Find $K$ most important features on client $k$ denoted as $\mathcal{F}_{important, k} \subseteq \mathcal{F}_{k}$\\ \Indm 
       \textbf{Hybrid}:\\ \Indp Find $K$ most important features across clients that share same feature space denoted as $\mathcal{F}_{important, k} \subseteq \mathcal{F}_{k}$. The set $\mathcal{F}_{important, k}$ will be the same across clients that share the same feature space. \\ \Indm 
   \For{each client $k=1,...,Q$}{
       $\mathcal{H}_k = [ \;]$\\
       \For{each $f \in \mathcal{F}_{important, k}$}{
            Homomorphic encrypt feature vector denote as $H(f)$\\
            $\mathcal{H}_k = \mathcal{H}_k \cup H(f)$
       }
       Send list of encrypted feature vectors on client $k$, $\mathcal{H}_k$, to server
    }
    Server executes CreateAndSendNonLocalCombinations($[\mathcal{H}_1,...,\mathcal{H}_Q]$)\\
    \For{each client $k=1,...,Q$}{
    Client $k$ decrypts new features $F_{i,j}=\text{Dec}(H_{i,j})$ and appends features to local data matrix $X_k$\\
    }
    \textbf{Vertical}:\\ \Indp Each client $k=1,...,Q$ selects best newly engineered features, and update local data matrix $X_k$\\ \Indm 
    \textbf{Hybrid}:\\ \Indp Each group of clients that share the same feature space coordinate and select best newly engineered features, and update local data matrix. Clients that share the same feature space must select the same feature indices to keep features consistent across horizontal splits.\\ \Indm 
}
\end{algorithm}

\begin{algorithm}[htbp]
\label{createlocal}
\caption{CreateLocalCombinations()}
    \textbf{Input}: local data matrix $X_k$ where we denote the $i$-th column as $X_k[i]$, set of candidate bivariate transformations $\mathcal{B}_{local}$, denote $\mathcal{X}_k$ as the set of of all columns of $X_k$ (so we can define set operations)\\
    $F_{out} = [\;]$\\
    \For{each $(X_k[i],X_k[j]) \in \mathcal{X}_k \times \mathcal{X}_k$}{
        \For{each candidate bivariateFunction() in $\mathcal{B}_{local}$}{
            $F_{out} = F_{out} \cup \text{bivariateFunction}(X_k[i],X_k[j])$
        }
    }
    Return $F_{out}$
\end{algorithm}

\begin{algorithm}[htbp]
\label{createnonlocal}
\caption{CreateAndSendNonLocalCombinations()}
    \textbf{Input}: Lists of encrypted feature vectors for each client denoted as $\mathcal{H}_k$, CKKS Homomorphic encryption function $H(\cdot)$, $\text{Lap}(0|b)^{d}$ represents a vector of dimension $d$ with independent values each drawn from a Laplace distribution with scale $b$, set of candidate bivariate transformations $\mathcal{B}_{nonlocal}$, define the set of client groups that share same sample space $\mathcal{G}_s$ (in vertical case this is a single group that contains all clients), define the set of client groups that share same feature space $\mathcal{G}_f$ (in vertical case each group will have a single client) \\
    $\text{IdxTracker}=\{\}$ \tcp{Tracker object that maps feature index pair key $(i,j)$ to feature client group in $\mathcal{G}_f$}
    \For{each client group $G$ in $\mathcal{G}_s$}{
        $\mathcal{H} = \{\}$\\
        \For{each client $k$ in $G$}{
            $\mathcal{H} = \mathcal{H} \cup \mathcal{H}_k$
        }
        \For{each $(H_i,H_j) \in \mathcal{H} \times \mathcal{H}$}{
            \For{each candidate bivariateFunction() in $\mathcal{B}_{nonlocal}$}{
                $H_{i,j} = \text{bivariateFunction}(H_i,H_j)+ H(\text{Lap}(0|b)^{N_k})$ \tcp{inject encrypted Laplace noise}
                \textbf{Vertical:}\\
                \Indp Select client that owns feature $i$ or feature $j$ at random with equal probability of selecting between the two. Send $H_{i,j}$ back to selected client \\ \Indm
                \textbf{Hybrid:}\\
                \Indp\eIf{$(i,j)$ in keys of $\text{IdxTracker}$}
                {   
                    Find client $c$ at intersection of $\text{IdxTracker}[(i,j)]$ and current sample client group $G$\\
                    Send $H_{i,j}$ to $c$
                }
                {
                    Select client that owns feature $i$ or feature $j$ at random with equal probability and send $H_{i,j}$ to selected client. Find feature client group $G_c \in \mathcal{G}_f$ that selected client belongs to\\
                    Update $\text{IdxTracker}[(i,j)] = G_c$
                }
                \Indm
            }
        }
    }
\end{algorithm}

\section{Experiments}
\label{sec:experiments}
We evaluate the performance of our algorithms in the horizontal, vertical, and hybrid FL settings. We test on the same eight datasets evaluated in \cite{iife}, which consist of four regression and four classification datasets. Details of these datasets are available in the Appendix Section \ref{app:setup}. We use L2-regularized linear models for these experiments; logistic regression for classification problems and ridge regression for regression problems. We perform random search hyperparameter tuning before the AutoFE process and after the AutoFE process. For linear models, the only parameter tuned is the regularization constant. The hyperparameter tuning evaluations are performed in an FL fashion. Following the approach in \cite{iife}, our test metric is F1-micro for classification problems and 1 - (relative absolute error) for regression problems. For the linear models we use one-hot encoding of the categorical variables and standard scaling of the features when training and performing evaluations. We define $Q$ as the number of clients.

\subsection{Horizontal FL Setting}
\subsubsection{Algorithm Performance}
\label{results:horizontal}
For the horizontal FL setting, we explore four different scenarios of how the data is gathered across clients. These scenarios are IID, IID with unequal sample sizes across clients, non-IID by target distribution, and non-IID by feature distribution. The non-IID by feature distribution is formed by performing $K$-means clustering where we set $K=Q$. Then each cluster is allocated to a particular client. The non-IID by target distribution is formed differently for regression and classification problems. For regression problems, we find the minimum and maximum target variable which we denote as $y_{min}$ and $y_{max}$, respectively. We then form a linear partition of $Q$ blocks each with length $\frac{y_{max}-y_{min}}{Q}$, where each block corresponds to a particular client. We assign each sample to a particular block based on the value of the target variable of that sample. Classification problems follow a similar technique, but a few samples of the other classes are included on each client to avoid some clients only having one class. These two schemes result in highly non-IID distributions. More details of how we perform these splits in our experiments is detailed in Section \ref{app:setup} of the Appendix. 

\begin{figure*}[htbp]
\centerline{\includegraphics[width=.99\textwidth]{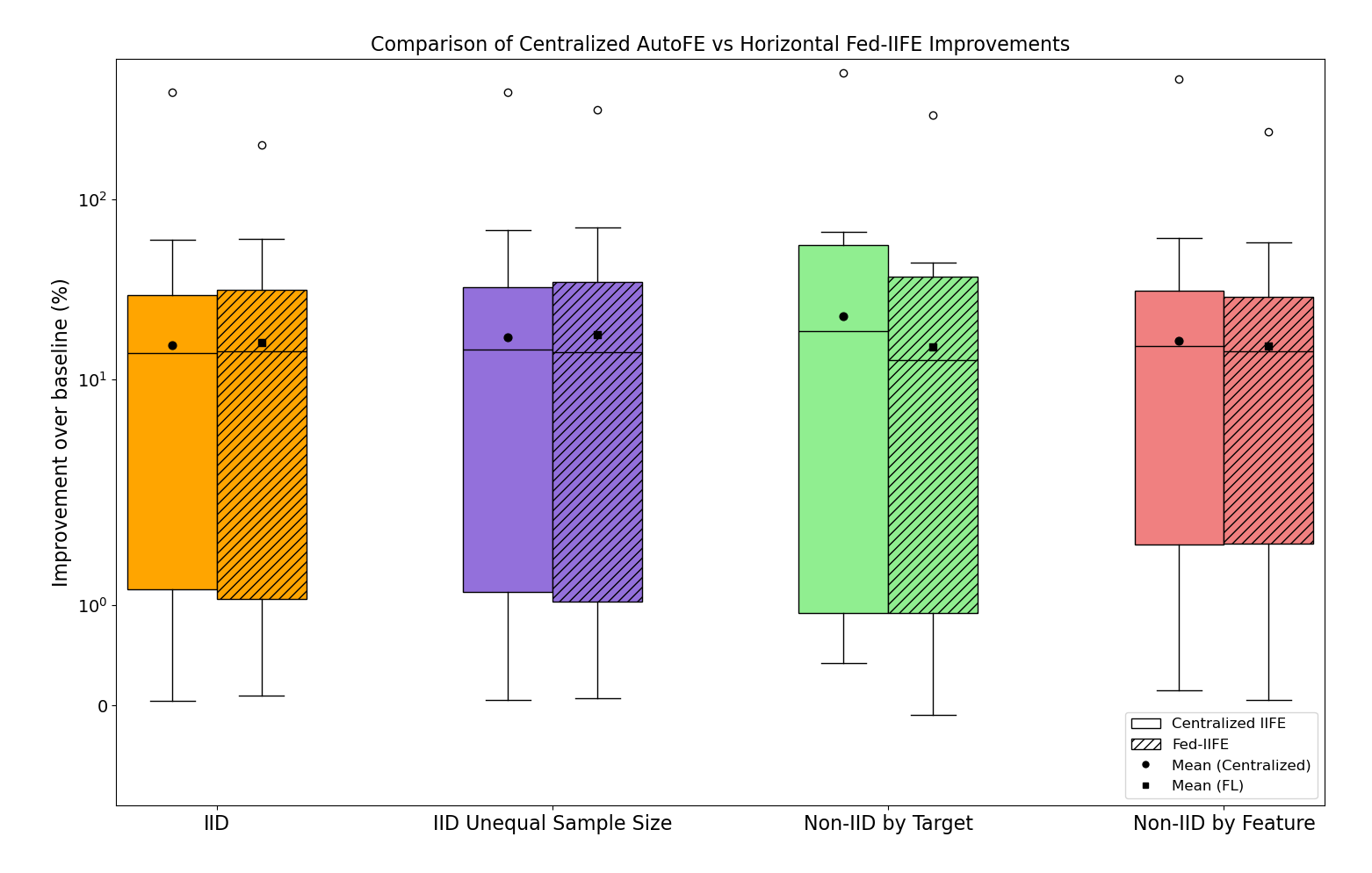}}
\caption{Fed-IIFE Results. Omitted outlier dataset OpenML586 which had extremely high improvement over baseline. All four settings have different baselines, because baselines and centralized scores are also computed with FL training.}
\label{horizontal_results}
\end{figure*}

We evaluate the base score without any engineered features, the case where AutoFE is performed centrally to create the engineered features, and the federated case where we use our algorithms to create the engineered features. All final training and evaluations are performed in the particular FL setting. Figure \ref{horizontal_results} shows the performance of Fed-IIFE in each of the horizontal FL scenarios compared to the case where the engineered features are built in the traditional, fully-centralized case. The metric being plotted is the percentage improvement over the baseline case where only the original features are used. The results show that for the IID cases, the performance of Fed-IIFE is nearly identical to the centralized case, and there is only a slight degradation in the non-IID cases. These results demonstrate that Fed-IIFE performs exceptionally well even in highly non-IID scenarios. To the best of our knowledge there are no horizontal AutoFE algorithms that exist to compare with our algorithm, and thus we are focused on comparisons with the top-performing centralized AutoFE algorithms.

Overall, we see improvements over the baseline raw features of 15.11\% for the difficult non-IID by target setting averaged across all datasets (excluding extreme positive outlier dataset OpenML586). This demonstrates that Fed-IIFE can generate very useful engineered features even in difficult non-IID horizontal FL settings.

\subsubsection{Verification of Federated Interaction Information}
\label{results:fed-ii-validation}
Computing interaction information values to predict the synergy of a feature pair with the target is a key step in IIFE. In \cite{iife}, the viability of interaction information was evaluated using a synthetic data experiment. The basic idea is to create a synthetic target as some function of two features. We would then expect the interaction information of the selected pair of features and the synthetic label to be high compared to other feature pairs. If we repeat this across many feature pairs in the dataset, in the optimal case, the true feature pair would be ranked 0 in terms of ranking by interaction information for every tested feature pair.

\begin{figure*}[htbp]
\centerline{\includegraphics[width=.99\textwidth]{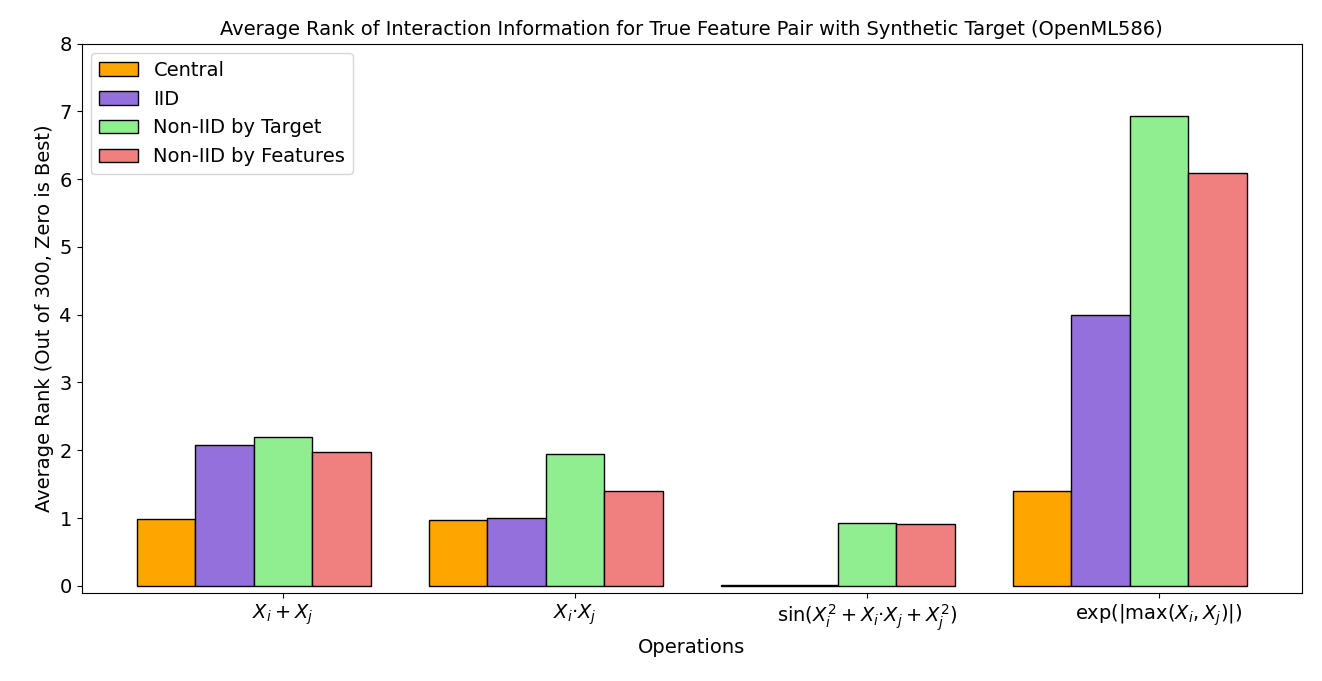}}
\caption{Synthetic data verification of federated interaction information}
\label{synthetic_experiment}
\end{figure*}

We repeat this experiment but for the case of federated interaction information where each client individually computes interaction information values with their available samples and then the server averages these interaction information values. We perform this on the OpenML586 dataset which has 25 features and thus 300 combinations of feature pairs. Figure \ref{synthetic_experiment} demonstrates that federated interaction information captures the synergy of pairs of features and the target variable very well, even for highly non-IID settings with very complicated synthetic functions.

\subsubsection{Similarity of Engineered Features between Centralized and FL AutoFE}
We are also interested in whether the important engineered features found in the centralized case are similar to the engineered features found with Fed-IIFE. In order to test this, we first find the top 25\% most important engineered features in the centralized case by permutation importance. For each of these most important centralized engineered features, we find the cosine distance (1 - cosine similarity) with each of the Fed-IIFE engineered features and take the minimum of these distances. The mean of these minimum cosine distances gives a good measure of whether Fed-IIFE is able to find similar important engineered features to those found with centralized IIFE. 

\begin{figure*}[htbp]
\centerline{\includegraphics[width=.7\textwidth]{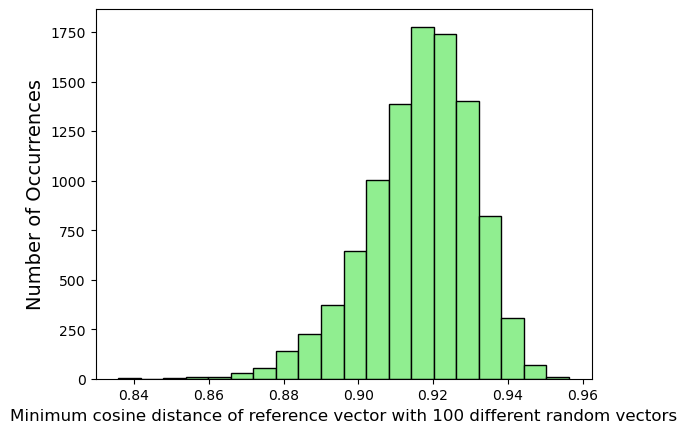}}
\caption{Cosine Distance Simulation for high-dimensional Random Vectors}
\label{cosine_simulation}
\end{figure*}

These cosine distances are difficult to interpret in isolation, as these distances are dependent on the dimensionality of the vectors. In very high dimensional space, random vectors tend to be orthogonal. We form a basic simulation, and for each trial we generate a random reference vector of dimension 900 (similar to the airfoil training set sample size). Then, for each reference vector we generate 100 more random vectors (this is a conservative number, the number of engineered features is usually smaller than this) and find the minimum cosine distance to the reference vector across all other random vectors. We show a histogram of these minimum cosine distances for 10,000 trials in Figure \ref{cosine_simulation}. It is clear that even when taking the minimum over a large number of random vectors, the minimum cosine distance tends to be a large number indicating that it is nearly orthogonal with the reference vector. 

In Table \ref{tab:feature_analysis}, we show the minimum cosine distances between the important centralized IIFE engineered features and the Fed-IIFE engineered features. These are averaged over 10 runs. We can see that for the IID case, the important centralized IIFE engineered features are almost exactly matched by Fed-IIFE. For the non-IID cases, the features are not quite as similar, but as shown in the random simulation in Figure \ref{cosine_simulation}, these cosine distances are still quite small in such a high-dimensional space and are significantly closer than random. Furthermore, there seems to be an inverse relationship with the minimum cosine distance and the relative test performance, which is defined as the Fed-IIFE test score divided by the centralized IIFE test score. 

\begin{table}[htbp]
    \centering
    \begin{tabular}{|c|c|c|}
    \hline
        Setting & Minimum cosine distance & Relative test performance \\ \hline
         IID & 0.0920 (0.0326) & 100.3\% \\ \hline
         Non-IID by target & 0.3001 (0.0648) & 86.8\% \\ \hline
         Non-IID by features & 0.2981 (0.0783) & 97.7\% \\ \hline
    \end{tabular}
    \caption{Minimum cosine distances in different horizontal FL scenarios for the Airfoil dataset. Averaged over 10 runs. Standard deviations shown in ().}
    \label{tab:feature_analysis}
\end{table}

\subsection{Vertical FL Setting}
\subsubsection{Algorithm Performance}
\label{sec:vertical_results}

\begin{figure*}[htbp]
\centerline{\includegraphics[width=.99\textwidth]{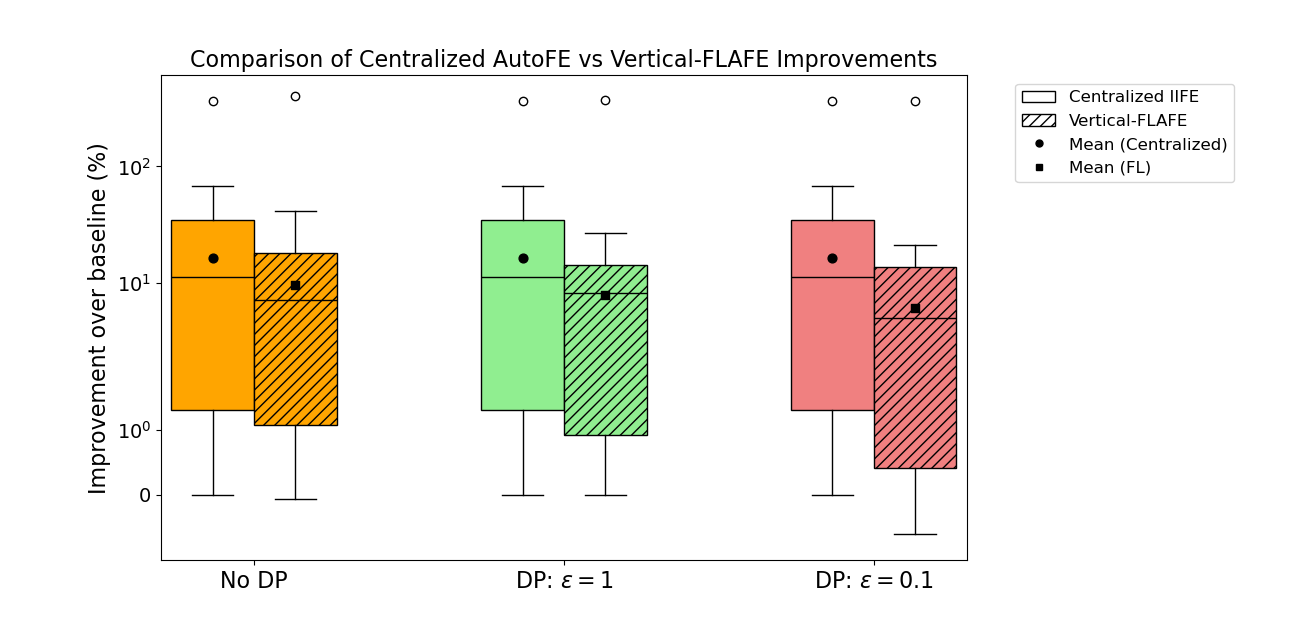}}
\caption{Vertical-FLAFE improvement over baseline original features for different levels of differential privacy compared with centralized IIFE. Means exclude OpenML586 outlier dataset that has extremely large improvements over baseline.}
\label{vertical_results}
\end{figure*}

For the Vertical FL AutoFE experiments, we create splits in the data by simply partitioning the feature space into equal segments. Due to each client having different features, the distributions of data across clients are non-IID by default. To protect client data from being reconstructed by the server or other clients, differential privacy is used when forming the feature combinations of features that belong on different clients as shown in Algorithm \ref{vertical-flafe}. We use additive Laplace noise which is $\epsilon$-differentially private \cite{laplace}. We run experiments with no differential privacy and with differential privacy with $\epsilon=1$ and $\epsilon=0.1$, which are relatively low values indicating high privacy. We perform homomorphic encryption using the CKKS scheme \cite{ckks} provided in the PyFHEL Python library \cite{pyfhel}. We use HyFDCA as the FL training algorithm as vertical FL is a special case of hybrid FL and performs well in this setting \cite{hyfdca}. Exact details of the experimental setup including differential privacy details are provided in Section \ref{app:setup} of the Appendix.

Figure \ref{vertical_results} demonstrates the performance of vertical-FLAFE compared to centralized IIFE, which is the best-performing centralized AutoFE algorithm. The centralized IIFE score is found by creating the engineered features centrally and then training the final model with HyFDCA in the vertical FL setting. We plot the improvement over the baseline (only original features) test score. We can see that there is a more obvious degradation in the performance for vertical FL than for horizontal FL with Fed-IIFE. This is due to the algorithmic limitations in the vertical FL case that were discussed in Section \ref{sec:vertical_algo}. However, even with these limitations, vertical-FLAFE shows significant improvements over the baseline score for the majority of datasets and even approaches the performance of centralized IIFE. We also see a decrease in performance as we move from no differential privacy to a high level of differential privacy with $\epsilon=0.1$. However, even with the very strict and secure $\epsilon=0.1$, we still see nice improvements over baseline raw features, indicating that informative engineered features are found by Vertical-FLAFE. To the best of our knowledge there are no vertical AutoFE works that are reproducible to compare with our algorithm (see discussion of one existing vertical AutoFE approach that is not reproducible in Section \ref{sec:related_work}), and thus we are focused on comparisons with the top-performing centralized AutoFE algorithms.

Table \ref{tab:effect_of_he} shows that homomorphic encryption is the clear bottleneck in Vertical-FLAFE. To obtain these results, we perform the full Vertical-FLAFE algorithm with and without CKKS homomorphic encryption. Removing homomorphic encryption is done for both the feature building stage as well as the model trainings which require homomorphic encryption. The runtimes shown include the time for hyperparameter tuning, Vertical-FLAFE, and training time of the final downstream model. It is clear from these results that homomorphic encryption significantly slows down the algorithm and is the key reason why a more sophisticated algorithm with many evaluated features is not feasible.

Overall, we see improvements over the baseline raw features of 6.09\% for the most secure differential privacy setting of $\epsilon=0.1$ averaged across all datasets (excluding extreme positive outlier dataset OpenML586). This demonstrates that Vertical-FLAFE can generate fairly useful engineered features even in the difficult vertical setting that requires very restrictive homomorphic encryption and differential privacy.

\begin{table}[htbp]
    \centering
    \begin{tabular}{|c|c|c|}
        \hline
        Dataset & Homomorphic Encryption & Total Vertical-FLAFE Runtime (s) \\
        \hline
        \multirow{2}{*}{OpenML586 (Regression)} & Yes & 2709.89 \\
                                   & No & 27.53 \\
        \hline
        \multirow{2}{*}{Jungle Chess (Classification)} & Yes & 5662.83 \\
                                      & No & 392.74 \\
        \hline
    \end{tabular}
    \caption{Effect of homomorphic encryption on average runtimes of Vertical-FLAFE}
    \label{tab:effect_of_he}
\end{table}

\subsection{Hybrid FL Setting}
\subsubsection{Algorithm Performance}

\begin{figure*}[htbp]
\centerline{\includegraphics[width=.99\textwidth]{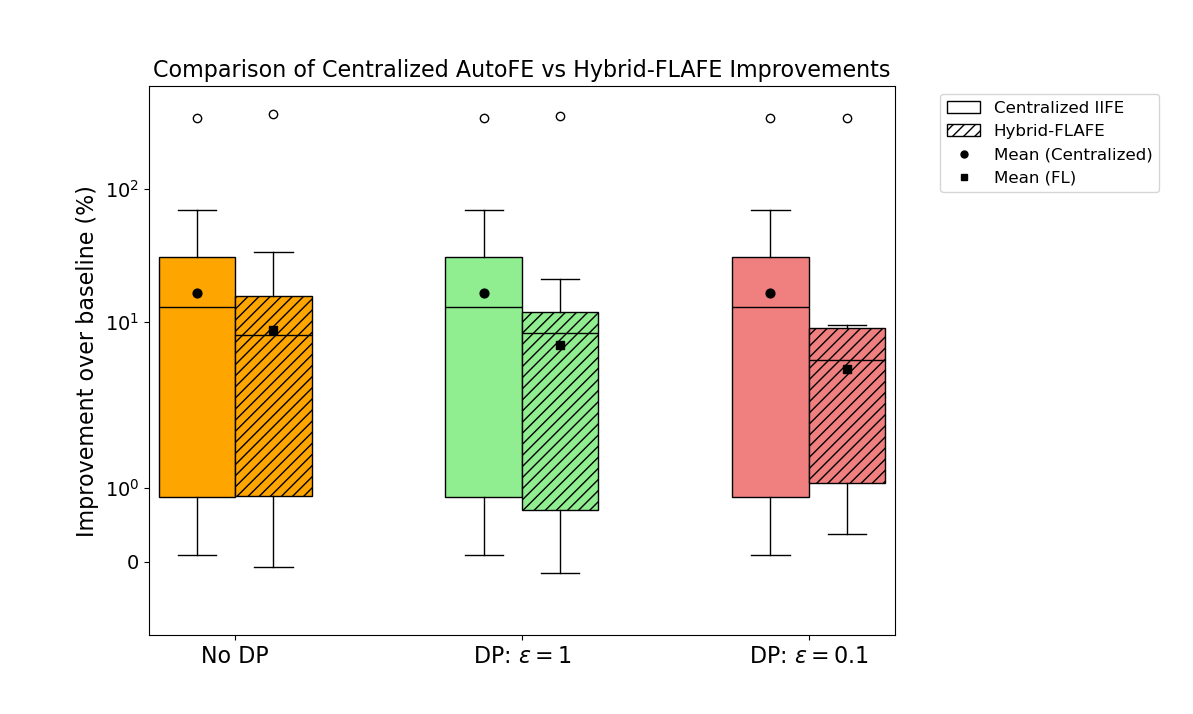}}
\caption{Hybrid-FLAFE improvement over baseline original features for different levels of differential privacy compared with centralized IIFE. Means exclude OpenML586 outlier dataset that has extremely large improvements over baseline.}
\label{hybrid_results}
\end{figure*}

 The experimental setup for the hybrid setting is very similar to the vertical setting as described in Section \ref{sec:vertical_results}. The main difference is that we use Hybrid-FLAFE to build new features and partition the data by the horizontal and vertical axes. Exact details of the experimental setup for the hybrid setting are provided in Section \ref{app:setup} of the Appendix.

The main takeaways for Hybrid-FLAFE are very similar to Vertical-FLAFE. We see an overall decrease in performance as we increase the strictness of differential privacy; however, even in the most strict setting the improvements over baseline are still very positive indicating that Hybrid-FLAFE finds useful features. In general, we see the expected loss in performance as we increase the strictness of differential privacy. As with Vertical-FLAFE, the main computational bottleneck in Hybrid-FLAFE is homomorphic encryption. To the best of our knowledge there are no hybrid AutoFE algorithms that exist to compare with our algorithm, and thus we are focused on comparisons with the top-performing centralized AutoFE algorithms.

Overall, we see improvements over the baseline raw features of 4.38\% for the most secure differential privacy setting of $\epsilon=0.1$ averaged across all datasets (excluding extreme positive outlier dataset OpenML586). This demonstrates that Hybrid-FLAFE can generate fairly useful engineered features even in the difficult hybrid FL setting that requires very restrictive homomorphic encryption and differential privacy.

\section{Conclusions}
We contribute three AutoFE algorithms that operate in the three key federated learning settings: horizontal, vertical, and hybrid. We provide experimental evidence that Horizontal-FLAFE performs similarly to performing the AutoFE process in a centralized manner. We also validate federated interaction information as a good measure of synergy between feature pairs and the target variable. We also find that the engineered features are quite similar between centralized IIFE and Fed-IIFE, even in strongly non-IID settings. Finally, we demonstrate that Vertical-FLAFE and Hybrid-FLAFE show strong improvements over baseline scores, even though the need for homomorphic encryption and differential privacy limits these algorithms more than the horizontal setting.

\bibsep 12pt
\nocite{}
\bibliographystyle{apalike} 
\bibliography{main.bib}

\newpage
\section{Appendix}
\label{appendix}

\subsection{Full Experimental Results}
\label{app:results}
\subsubsection{Full Performance Results}

We show the full scores for all three algorithms in all of the studied settings in Tables \ref{tab:horizontal_autofe_results}, \ref{tab:vertical_autofe_results}, and \ref{tab:hybrid_autofe_results}. The base score is the test score with only the original features, but these still include training in the FL setting specified. The centralized IIFE score is found by creating the engineered features in a centralized manner, but still train and evaluate the final model in a federated setting using FL training algorithms. This is because we are interested in the effect of FL on the engineered features being built, not on the performance of the final FL training algorithm. All fo these results are found over 10 runs with different random seeds.

\begin{table}[htbp]
\centering
\begin{tabular}{|l|l|c|c|c|}
\hline
\textbf{Dataset-Task} & \textbf{Configuration} & \textbf{Base score} & \textbf{Centralized IIFE } & \textbf{Fed-IIFE} \\
\hline

\multirow{4}{*}{Airfoil - Lasso} 
& IID & 0.3253 (0.0271) & 0.5180 (0.0402) & 0.5196 (0.0390) \\
& IID unequal sample size & 0.3245 (0.0270) & 0.5422 (0.0324) & 0.5501 (0.0376) \\
& Non-IID by target & 0.3240 (0.0268) & 0.5378 (0.0380) & 0.4670 (0.0417) \\
& Non-IID by feature & 0.3229 (0.0316) & 0.5200 (0.0296) & 0.5083 (0.0420) \\
\hline

\multirow{4}{*}{Jungle Chess - LR}
& IID & 0.6810 (0.0034) & 0.7707 (0.0115) & 0.7788 (0.0118) \\
& IID unequal sample size & 0.6812 (0.0034) & 0.7741 (0.0089) & 0.7783 (0.0070) \\
& Non-IID by target & 0.5204 (0.0031) & 0.7899 (0.0103) & 0.7018 (0.0186) \\
& Non-IID by feature & 0.6751 (0.0054) & 0.7865 (0.0095) & 0.7691 (0.0097) \\
\hline

\multirow{4}{*}{OpenML 586 - Lasso}
& IID & 0.1348 (0.0223) & 0.6636 (0.0711) & 0.4031 (0.1073) \\
& IID unequal sample size & 0.1300 (0.0235) & 0.6410 (0.0543) & 0.5356 (0.1097) \\
& Non-IID by target & 0.1152 (0.0272) & 0.6934 (0.0463) & 0.4526 (0.0906) \\
& Non-IID by feature & 0.1252 (0.0289) & 0.7071 (0.0301) & 0.4192 (0.0762) \\
\hline

\multirow{4}{*}{Bikeshare - Lasso}
& IID & 0.2771 (0.0413) & 0.3176 (0.0495) & 0.3161 (0.0289) \\
& IID unequal sample size & 0.2778 (0.0408) & 0.3208 (0.0365) & 0.3171 (0.0382) \\
& Non-IID by target & 0.2754 (0.0419) & 0.3235 (0.0207) & 0.3219 (0.0351) \\
& Non-IID by feature  & 0.2769 (0.0395) & 0.3158 (0.0355) & 0.3172 (0.0367) \\
\hline

\multirow{4}{*}{JM1 - LR}
& IID & 0.8127 (0.0059) & 0.8131 (0.0063) & 0.8135 (0.0056) \\
& IID unequal sample size & 0.8127 (0.0061) & 0.8132 (0.0068) & 0.8133 (0.0056) \\
& Non-IID by target & 0.8102 (0.0060) & 0.8136 (0.0072) & 0.8095 (0.0056) \\
& Non-IID by feature & 0.8127 (0.0068) & 0.8139 (0.0073) & 0.8132 (0.0069) \\
\hline

\multirow{4}{*}{Winequalityred - LR}
& IID & 0.5938 (0.0248) & 0.6016 (0.0261) & 0.6003 (0.0232) \\
& IID unequal sample size & 0.5938 (0.0257) & 0.6013 (0.0216) & 0.5997 (0.0254) \\
& Non-IID by target & 0.5884 (0.0284) & 0.5947 (0.0303) & 0.5956 (0.0256) \\
& Non-IID by feature & 0.5953 (0.0239) & 0.5984 (0.0252) & 0.6000 (0.0277) \\
\hline

\multirow{4}{*}{Cal Housing - Lasso}
& IID & 0.4160 (0.0086) & 0.4970 (0.0219) & 0.5078 (0.0092) \\
& IID unequal sample size & 0.4157 (0.0089) & 0.5032 (0.0097) & 0.5123 (0.0121) \\
& Non-IID by target & 0.4212 (0.0088) & 0.5042 (0.0098) & 0.4579 (0.0712) \\
& Non-IID by feature & 0.4164 (0.0090) & 0.5029 (0.0117) & 0.4951 (0.0119) \\
\hline

\multirow{4}{*}{Credit Default - LR}
& IID & 0.8094 (0.0045) & 0.8152 (0.0031) & 0.8173 (0.0034) \\
& IID unequal sample size & 0.8095 (0.0044) & 0.8157 (0.0033) & 0.8180 (0.0031) \\
& Non-IID by target & 0.8107 (0.0037) & 0.8147 (0.0042) & 0.8111 (0.0055) \\
& Non-IID by feature & 0.8006 (0.0145) & 0.8164 (0.0036) & 0.8157 (0.0042) \\
\hline

\end{tabular}
\caption{Horizontal FL AutoFE: Performance of Fed-IIFE under various federated configurations across datasets. Standard deviations shown in parentheses ().}
\label{tab:horizontal_autofe_results}
\end{table}

\begin{table}[htbp]
\centering
\begin{tabular}{|l|l|c|c|c|}
\hline
\textbf{Dataset-Task} & \textbf{Configuration} & \textbf{Base score} & \textbf{Centralized IIFE} & \textbf{Vertical-FLAFE} \\
\hline

\multirow{3}{*}{Airfoil - Lasso}
& No DP & 0.3301 (0.0449) & 0.5531 (0.0508) & 0.4665 (0.0655) \\
& DP - $\epsilon$=1 & 0.3301 (0.0449) & 0.5531 (0.0508) & 0.4192 (0.0730) \\
& DP - $\epsilon$=0.1 & 0.3301 (0.0449) & 0.5531 (0.0508) & 0.4005 (0.0694) \\
\hline

\multirow{3}{*}{Jungle Chess - LR}
& No DP & 0.6788 (0.0027) & 0.7203 (0.0469) & 0.7460 (0.0093) \\
& DP - $\epsilon$=1 & 0.6788 (0.0027) & 0.7203 (0.0469) & 0.7382 (0.0048) \\
& DP - $\epsilon$=0.1 & 0.6788 (0.0027) & 0.7203 (0.0469) & 0.7392 (0.0044) \\
\hline

\multirow{3}{*}{OpenML 586 - Lasso}
& No DP & 0.1162 (0.0427) & 0.5349 (0.1879) & 0.5841 (0.0499) \\
& DP - $\epsilon$=1 & 0.1162 (0.0427) & 0.5349 (0.1879) & 0.5441 (0.0400) \\
& DP - $\epsilon$=0.1 & 0.1162 (0.0427) & 0.5349 (0.1879) & 0.5407 (0.0329) \\
\hline

\multirow{3}{*}{Bikeshare - Lasso}
& No DP & 0.2409 (0.0561) & 0.2812 (0.0385) & 0.2520 (0.0509) \\
& DP - $\epsilon$=1 & 0.2409 (0.0561) & 0.2812 (0.0385) & 0.2600 (0.0501) \\
& DP - $\epsilon$=0.1 & 0.2409 (0.0561) & 0.2812 (0.0385) & 0.2678 (0.0571) \\
\hline

\multirow{3}{*}{JM1 - LR}
& No DP & 0.8140 (0.0092) & 0.8140 (0.0088) & 0.8134 (0.0095) \\
& DP - $\epsilon$=1 & 0.8140 (0.0092) & 0.8140 (0.0088) & 0.8140 (0.0087) \\
& DP - $\epsilon$=0.1 & 0.8140 (0.0092) & 0.8140 (0.0088) & 0.8140 (0.0088) \\
\hline

\multirow{3}{*}{Winequalityred - LR}
& No DP & 0.5825 (0.0446) & 0.5913 (0.0394) & 0.5875 (0.0427) \\
& DP - $\epsilon$=1 & 0.5825 (0.0446) & 0.5913 (0.0394) & 0.5884 (0.0406) \\
& DP - $\epsilon$=0.1 & 0.5825 (0.0446) & 0.5913 (0.0394) & 0.5900 (0.0414) \\
\hline

\multirow{3}{*}{Cal Housing - Lasso}
& No DP & 0.3987 (0.0096) & 0.4931 (0.0083) & 0.4409 (0.0199) \\
& DP - $\epsilon$=1 & 0.3987 (0.0096) & 0.4931 (0.0083) & 0.4393 (0.0406) \\
& DP - $\epsilon$=0.1 & 0.3987 (0.0096) & 0.4931 (0.0083) & 0.3963 (0.1532) \\
\hline

\multirow{3}{*}{Credit Default - LR}
& No DP & 0.8069 (0.0078) & 0.8124 (0.0083) & 0.8162 (0.0064) \\
& DP - $\epsilon$=1 & 0.8070 (0.0078) & 0.8124 (0.0083) & 0.8124 (0.0092) \\
& DP - $\epsilon$=0.1 & 0.8070 (0.0078) & 0.8124 (0.0083) & 0.8115 (0.0095) \\
\hline

\end{tabular}
\caption{Vertical FL AutoFE: Performance of Vertical-FLAFE under various privacy configurations across datasets. Standard deviations shown in parentheses ().}
\label{tab:vertical_autofe_results}
\end{table}

\begin{table}[htbp]
\centering
\begin{tabular}{|l|l|c|c|c|}
\hline
\textbf{Dataset-Task} & \textbf{Configuration} & \textbf{Base score} & \textbf{Centralized IIFE} & \textbf{Hybrid-FLAFE} \\
\hline

\multirow{3}{*}{Airfoil - Lasso}
& No DP & 0.3341 (0.0439) & 0.5668 (0.0410) & 0.4451 (0.0374) \\
& DP - $\epsilon$=1 & 0.3341 (0.0439) & 0.5668 (0.0410) & 0.4039 (0.0593) \\
& DP - $\epsilon$=0.1 & 0.3341 (0.0439) & 0.5668 (0.0410) & 0.3622 (0.0410) \\
\hline

\multirow{3}{*}{Jungle Chess - LR}
& No DP & 0.6784 (0.0031) & 0.7563 (0.0298) & 0.7238 (0.0086) \\
& DP - $\epsilon$=1 & 0.6784 (0.0031) & 0.7563 (0.0298) & 0.7311 (0.0100) \\
& DP - $\epsilon$=0.1 & 0.6784 (0.0031) & 0.7563 (0.0298) & 0.7387 (0.0040) \\
\hline

\multirow{3}{*}{OpenML 586 - Lasso}
& No DP & 0.1220 (0.0383) & 0.5398 (0.1721) & 0.5728 (0.0439) \\
& DP - $\epsilon$=1 & 0.1220 (0.0383) & 0.5398 (0.1721) & 0.5546 (0.0419) \\
& DP - $\epsilon$=0.1 & 0.1220 (0.0383) & 0.5398 (0.1721) & 0.5395 (0.0484) \\
\hline

\multirow{3}{*}{Bikeshare - Lasso}
& No DP & 0.2515 (0.0554) & 0.2869 (0.0480) & 0.2758 (0.0407) \\
& DP - $\epsilon$=1 & 0.2515 (0.0554) & 0.2869 (0.0480) & 0.2729 (0.0847) \\
& DP - $\epsilon$=0.1 & 0.2515 (0.0554) & 0.2869 (0.0480) & 0.2563 (0.1186) \\
\hline

\multirow{3}{*}{JM1 - LR}
& No DP & 0.8085 (0.0101) & 0.8092 (0.0099) & 0.8079 (0.0086) \\
& DP - $\epsilon$=1 & 0.8085 (0.0101) & 0.8092 (0.0099) & 0.8073 (0.0132) \\
& DP - $\epsilon$=0.1 & 0.8085 (0.0101) & 0.8092 (0.0099) & 0.8115 (0.0096) \\
\hline

\multirow{3}{*}{Winequalityred - LR}
& No DP & 0.5800 (0.0424) & 0.5859 (0.0372) & 0.5861 (0.0403) \\
& DP - $\epsilon$=1 & 0.5800 (0.0424) & 0.5859 (0.0372) & 0.5847 (0.0377) \\
& DP - $\epsilon$=0.1 & 0.5800 (0.0424) & 0.5859 (0.0372) & 0.5875 (0.0437) \\
\hline

\multirow{3}{*}{Cal Housing - Lasso}
& No DP & 0.4125 (0.0088) & 0.4863 (0.0094) & 0.4502 (0.0242) \\
& DP - $\epsilon$=1 & 0.4125 (0.0088) & 0.4863 (0.0094) & 0.4486 (0.0220) \\
& DP - $\epsilon$=0.1 & 0.4125 (0.0088) & 0.4863 (0.0094) & 0.4511 (0.0224) \\
\hline

\multirow{3}{*}{Credit Default - LR}
& No DP & 0.8082 (0.0084) & 0.8117 (0.0098) & 0.8117 (0.0103) \\
& DP - $\epsilon$=1 & 0.8082 (0.0084) & 0.8117 (0.0098) & 0.8113 (0.0090) \\
& DP - $\epsilon$=0.1 & 0.8082 (0.0084) & 0.8117 (0.0098) & 0.8114 (0.0093) \\
\hline

\end{tabular}
\caption{Hybrid FL AutoFE: Performance of Hybrid-FLAFE under various privacy configurations across datasets. Standard deviations shown in parentheses ().}
\label{tab:hybrid_autofe_results}
\end{table}

Now we show the runtimes for all three algorithms in all of the different FL settings in Tables \ref{tab:horizontal_runtime}, \ref{tab:vertical_runtime}, and \ref{tab:hybrid_runtime}. These runtimes include the time for hyperparameter tuning before and after AutoFE, the time for the AutoFE process, and the time it takes to train and evaluate the final model on the final engineered features.

\begin{table}[htbp]
\centering
\begin{tabular}{|l|l|c|}
\hline
\textbf{Dataset-Task} & \textbf{Configuration} & \textbf{Runtime (s)} \\
\hline

\multirow{4}{*}{Airfoil - Lasso}
& IID & 1256.95 (128.82) \\
& IID unequal sample size & 1620.65 (188.58) \\
& Non-IID by Target & 1331.49 (128.93) \\
& Non-IID by Features & 1332.80 (160.78) \\
\hline

\multirow{4}{*}{Jungle Chess - LR}
& IID & 2694.98 (469.53) \\
& IID unequal sample size & 1852.93 (230.47) \\
& Non-IID by Target & 3522.14 (402.10) \\
& Non-IID by Features & 2231.23 (521.45) \\
\hline

\multirow{4}{*}{OpenML 586 - Lasso}
& IID & 1799.12 (189.69) \\
& IID unequal sample size & 1906.94 (269.50) \\
& Non-IID by Target & 1526.01 (90.39) \\
& Non-IID by Features & 1554.15 (182.30) \\
\hline

\multirow{4}{*}{Bikeshare - Lasso}
& IID & 347.89 (53.91) \\
& IID unequal sample size & 388.23 (18.75) \\
& Non-IID by Target & 376.70 (20.74) \\
& Non-IID by Features & 365.97 (10.65) \\
\hline

\multirow{4}{*}{JM1 - LR}
& IID & 873.97 (131.62) \\
& IID unequal sample size & 962.48 (96.95) \\
& Non-IID by Target & 953.91 (124.82) \\
& Non-IID by Features & 816.93 (84.52) \\
\hline

\multirow{4}{*}{Winequalityred - LR}
& IID & 1270.63 (187.54) \\
& IID unequal sample size & 1243.84 (95.25) \\
& Non-IID by Target & 1159.42 (62.30) \\
& Non-IID by Features & 1014.52 (44.57) \\
\hline

\multirow{4}{*}{Cal Housing - Lasso}
& IID & 2716.02 (481.88) \\
& IID unequal sample size & 2310.15 (318.94) \\
& Non-IID by Target & 2047.88 (250.53) \\
& Non-IID by Features & 2224.19 (606.98) \\
\hline

\multirow{4}{*}{Credit Default - LR}
& IID & 2748.72 (502.65) \\
& IID unequal sample size & 1337.89 (190.00) \\
& Non-IID by Target & 1407.93 (257.01) \\
& Non-IID by Features & 1708.13 (371.65) \\
\hline

\end{tabular}
\caption{Runtime (in seconds) for Fed-IIFE across dataset-tasks. Standard deviations shown in parentheses ().}
\label{tab:horizontal_runtime}
\end{table}

\begin{table}[htbp]
\centering
\begin{tabular}{|l|l|c|}
\hline
\textbf{Dataset-Task} & \textbf{Configuration} & \textbf{Runtime (s)} \\
\hline

\multirow{3}{*}{Airfoil - Lasso}
& No DP & 8010.51 (2676.15) \\
& DP - $\epsilon$=1 & 7891.40 (2677.85) \\
& DP - $\epsilon$=0.1 & 7954.09 (2677.72) \\
\hline

\multirow{3}{*}{Jungle Chess - LR}
& No DP & 5851.13 (148.79) \\
& DP - $\epsilon$=1 & 5730.09 (149.48) \\
& DP - $\epsilon$=0.1 & 5662.83 (150.01) \\
\hline

\multirow{3}{*}{OpenML 586 - Lasso}
& No DP & 2740.77 (325.77) \\
& DP - $\epsilon$=1 & 2542.45 (323.37) \\
& DP - $\epsilon$=0.1 & 2709.89 (323.46) \\
\hline

\multirow{3}{*}{Bikeshare - Lasso}
& No DP & 13506.02 (1647.82) \\
& DP - $\epsilon$=1 & 13320.24 (1652.49) \\
& DP - $\epsilon$=0.1 & 13219.58 (1650.02) \\
\hline

\multirow{3}{*}{JM1 - LR}
& No DP & 1401.50 (67.45) \\
& DP - $\epsilon$=1 & 1417.96 (69.57) \\
& DP - $\epsilon$=0.1 & 1410.98 (69.29) \\
\hline

\multirow{3}{*}{Winequalityred - LR}
& No DP & 7853.98 (258.11) \\
& DP - $\epsilon$=1 & 7829.45 (258.72) \\
& DP - $\epsilon$=0.1 & 7808.70 (258.29) \\
\hline

\multirow{3}{*}{Cal Housing - Lasso}
& No DP & 3819.72 (153.63) \\
& DP - $\epsilon$=1 & 4254.30 (147.73) \\
& DP - $\epsilon$=0.1 & 3960.38 (156.96) \\
\hline

\multirow{3}{*}{Credit Default - LR}
& No DP & 1536.93 (75.48) \\
& DP - $\epsilon$=1 & 1524.85 (74.38) \\
& DP - $\epsilon$=0.1 & 1512.56 (74.69) \\
\hline

\end{tabular}
\caption{Runtime (in seconds) for Vertical-FLAFE across dataset-tasks. Standard deviations shown in parentheses ().}
\label{tab:vertical_runtime}
\end{table}

\begin{table}[htbp]
\centering
\begin{tabular}{|l|l|c|}
\hline
\textbf{Dataset-Task} & \textbf{Configuration} & \textbf{Runtime (s)} \\
\hline

\multirow{3}{*}{Airfoil - Lasso}
& No DP & 4490.49 (1147.88) \\
& DP - $\epsilon$=1 & 4506.45 (1161.52) \\
& DP - $\epsilon$=0.1 & 4548.80 (1166.42) \\
\hline

\multirow{3}{*}{Jungle Chess - LR}
& No DP & 2378.76 (314.69) \\
& DP - $\epsilon$=1 & 2363.85 (313.07) \\
& DP - $\epsilon$=0.1 & 2426.13 (322.35) \\
\hline

\multirow{3}{*}{OpenML 586 - Lasso}
& No DP & 2963.62 (1032.64) \\
& DP - $\epsilon$=1 & 2959.94 (1029.22) \\
& DP - $\epsilon$=0.1 & 2958.74 (1027.83) \\
\hline

\multirow{3}{*}{Bikeshare - Lasso}
& No DP & 11332.88 (2676.87) \\
& DP - $\epsilon$=1 & 11378.57 (2646.54) \\
& DP - $\epsilon$=0.1 & 11273.01 (2687.18) \\
\hline

\multirow{3}{*}{JM1 - LR}
& No DP & 1194.71 (77.84) \\
& DP - $\epsilon$=1 & 1315.87 (175.39) \\
& DP - $\epsilon$=0.1 & 1445.21 (93.76) \\
\hline

\multirow{3}{*}{Winequalityred - LR}
& No DP & 11586.47 (1007.62) \\
& DP - $\epsilon$=1 & 11610.67 (1024.58) \\
& DP - $\epsilon$=0.1 & 11652.81 (1018.57) \\
\hline

\multirow{3}{*}{Cal Housing - Lasso}
& No DP & 2279.43 (326.12) \\
& DP - $\epsilon$=1 & 2695.69 (422.22) \\
& DP - $\epsilon$=0.1 & 2281.00 (327.93) \\
\hline

\multirow{3}{*}{Credit Default - LR}
& No DP & 998.50 (48.32) \\
& DP - $\epsilon$=1 & 1046.58 (85.04) \\
& DP - $\epsilon$=0.1 & 1050.40 (101.72) \\
\hline

\end{tabular}
\caption{Runtime (in seconds) for Hybrid-FLAFE across dataset-tasks. Standard deviations shown in parentheses ().}
\label{tab:hybrid_runtime}
\end{table}

\subsubsection{Fully Local Feature Building Performs Poorly for Non-IID}
\label{sec:fullyindependent}
In the early phases of our research in horizontal FL AutoFE, we first tested the approach of performing AutoFE separately on each client and then taking the union of these transformations. However, we quickly found that this suffers very poor performance in highly non-IID settings. This process does not incorporate any information from other clients during the feature building stage which results in bad engineered features in non-IID settings. This is shown in Table \ref{tab:independent_autofe}, where we see good performance for IID settings, but a severe degradation for non-IID settings. These results are shown for the case where we perform AutoFeat (a top-performing AutoFE method for linear models) separately on each client, find the union of all engineered features, and then apply the feature selection approach shown in Algorithm \ref{sec:fed-halving}. As we are typically concerned about good performance for non-IID settings when studying FL algorithms, it is clear that this approach is not satisfactory. That is why we must incorporate some information from all clients while building the features, as we do in Fed-IIFE.

\begin{table}[htbp]
    \centering
    \begin{tabular}{|c|c|c|} \hline
         Setting & Base Score & Test Score after AutoFE  \\\hline
         IID & 0.3253 (0.0271) & 0.5378 (0.0405)\\\hline
         IID Unequal Sample Size & 0.3245 (0.0270) & 0.5862 (0.0380) \\\hline
         Non-IID by Target & 0.3240 (0.0268) & 0.4185 (0.0526)\\\hline
         Non-IID by Features & 0.3229 (0.0316) & 0.2619 (0.2080)\\\hline
    \end{tabular}
    \caption{Performance of AutoFeat Run Separately on each Client then Server Takes Union of Engineered Features in Horizontal FL settings. For Airfoil Dataset averaged over 10 runs. Standard deviations in parantheses ().}
    \label{tab:independent_autofe}
\end{table}

\subsection{Experimental Setup Details}
\label{app:setup}
We give the full details of our experimental setup to aid in reproducibility. Table \ref{dataset_summary} gives details and references on the datasets that we test. Furthermore, for the bikeshare dataset, the last two features when added together give exactly the target. Therefore, we remove these final two features to make the dataset more than just a trivial exercise.

\begin{table}
\caption{Summary of Datasets}
\begin{center}
\begin{tabular}{|c|r|r|c|}
\hline
\textbf{Dataset} & \textbf{\textit{\# Samples}}& \textbf{\textit{\# Features}}& \textbf{\textit{Type}} 
\\
\hline
Airfoil \cite{airfoil} & 1,503 & 5 & Regression  \\
\hline
Credit Default \cite{credit_default} & 30,000 & 23 & Classification \\
\hline
Bikeshare \cite{bikeshare} & 17,389 & 13 &  Regression \\
\hline
Wine Quality - Red \cite{winequality-red} & 999 & 12 & Classification \\
\hline
California Housing \cite{california_housing} & 20,640 & 8 & Regression \\
\hline
OpenML 586 \cite{fri} & 1,000 & 25 & Regression \\
\hline
JM1 \cite{jm1} & 10,885 & 22 &  Classification \\
\hline
Jungle Chess \cite{jungle_chess} & 44,819 & 6 & Classification \\
\hline
\end{tabular}
\label{dataset_summary}
\end{center}
\end{table}

\subsubsection{Horizontal Setting}
We use FedAvg as the algorithm of choice for model trainings \cite{fedavg}. We choose the simple case of $E=1$ local iterations before averaging. We use the weighted averaging scheme where weights are the ratio of samples on the client compared to the total number of samples across all clients. For final test score evaluations we generally use 2500 total global iterations of FedAvg which results in stable convergence. For the following datasets we are able to use less iterations and still reach convergence: jungle chess with 500 iterations, credit default with 1000 iterations, and jm1 with 500 iterations. For regression, we use L2 regularized linear regression, and for classification we use L2 regularized multinomial logistic regression.

We tune hyperparameters before and after AutoFE. We use a randomized hyperparameter search where the regularization constant is drawn log-uniformly from the range $[0.0001, 1000]$. In general, we have 50 iterations of this random hyperparameter search. But for the large dataset jungle chess, we only do 25 iterations. For the hyperparameter tuning, the models are trained in a federated fashion and then evaluated on a validation set that is held by the server.

For each dataset and setting, we run 10 trials each with different random seeds. These random seeds control any randomness in the train-test splits as well as randomness in the algorithms. For all horizontal experiments we use 8 clients and the main body of the paper details how the non-IID splits are made. For the IID unequal sample size, we first randomly assign half of the clients to client one, then a quarter to client two, and so on. Then the final client gets any remaining samples. For operations that are typically done by clients in parallel, we simulate in serial but take the max runtime out of all clients to simulate the true runtimes in practice. The data is split randomly each trial to create a train set (60\%), a test set (20\%), and a validation set (20\%). The test set is used for final evaluations. The operations we allow for binary functions are multiplication, minimum, maximum, first element divided by the absolute value of the second element plus 1 and the reverse order of this. For the unary operations, we allow square, absolute value, square root of the absolute value, and sigmoid.

\subsubsection{Vertical and Hybrid Settings}
The experimental setup for the vertical and hybrid settings follows similarly to the horizontal case, but there are several different aspects which we list in this section. For the vertical partitions in the data, we simply split the space into equal-sized sections. For the vertical case with the jungle chess, jm1, and credit default datasets, we split the feature space into five sections where each client gets one of these blocks, and for the other datasets we use 3 partitions. For the hybrid case with the jungle chess, jm1, and credit default datasets, we split both the feature space and sample space into five partitions for a total of twenty-five clients, and for the other datasets we use 3 partitions in each axis for a total of nine clients.

We actually use homomorphic encryption in our algorithm and choose to use PyFHEL, which is a python library that utilizes the underlying compiled C code to quickly do these operations. We use CKKS encryption scheme with the following parameters: $n=2^{15}$ (this is the polynomial modulus degree), scale=$2^{30}$ (this controls the float to fixed point conversion), and qi\_sizes=$[60,30,30,30,60]$. These are recommended values in the PyFHEL documentation. For some datasets with too many samples to encrypt all rows in the same vector, we have to split the encryption into several sections. This is the case for the jungle chess, credit default, california housing, and bikeshare datasets. Due to the homormorphic encryption scheme only preserving addition and multiplication, we are limited to more basic operations. For binary operations we allow multiplication, the square of the first element multiplied by the second element (and reverse), and the product of the square of both elements.

For adding Laplace noise, we first clip each elements of the vector (after standardization) to be within the range $[-5,5]$. We can then compute the sensitivity of the bivariate function that results in the largest sensitivity. This is the multiplication of the square of the two elements. This results in a sensitivity of $2c^4$ where $c$ is the clip value. We then compute $b=\frac{2*c^4}{\epsilon}$ and use this $b$ as the scale of our Laplace noise. The server is able to add this encrypted noise because it has a public key that can be used to encrypt, but not decrypt. Another privacy consideration is whether we can enforce that all clients do, in fact, clip their features. This can be verified with an honest client that has access to the private key and that receives the combination from the server and ensures the combination is within the specified bounds.

For the feature selection setup, we use random forest impurity-based feature importance on each client separately, so each client determines the most important features that belong to that client. We find this works well enough and is much faster than a fully federated feature selection technique such as that shown in Section \ref{app:alternative_algos}. For the hybrid case, we simply average the feature importance values across clients that share the same feature space. We perform one iteration in Vertical-FLAFE and select the top 10 features on each client after the local combinations are formed and before global combinations between clients are formed. Then finally, the top 10 features on each client are kept after global combinations are formed. These final features are appended to the original features. Even though these generated features are generally more simple than horizontal and centralized AutoFE algorithms, the performance still improves significantly over baseline.

\subsection{Alternative Algorithms}
\label{app:alternative_algos}
\begin{algorithm}[htbp]
\label{sec:fed-halving}
\caption{Fed-Halving: Federated Feature Selection}
\textbf{Input}: Clients $k=1,2,...,Q$, Local data matrix $X_k\in \mathbb{R}^{N_k\times M}$ and targets $y_k\in \mathbb{R}^{N_k}$ belonging to client $k$, denote the list of features as $\mathcal{F}$,\\
\textbf{Algorithm Parameters}: Number of sparsity-refinement iterations $I$, Number of halving iterations $H$, number of starting candidates for $h$-th iteration $\mathcal{N}_h$, halving ratio $\gamma$, number of levels in $h$-th halving iteration $\mathcal{L}_h$, number of communications in level $l$ of $h$-th halving iteration $\mathcal{C}_{h,l}$, Number of outermost server iterations $J$, number of sparsities to select each iteration $s$\\
$s_a=0$, $s_b=1$\\
\textbf{Server Executes:}\\
\For{each sparsity-refinement iteration $i=1,...,I$}{
    $\mathbb{L} = \{\;\}$\\
    \For{each halving iteration $h=1,2,...,H$}{
        $\mathbb{L}$ = $\mathbb{L}\; \cup$ FedSuccessiveHalving($\mathcal{F}$, $\mathcal{N}_h$, $\gamma$, $\mathcal{L}_h$, $\{\mathcal{C}_{h,l} \; \forall \; l \in 1,...,\mathcal{L}_h\}$, $s_a$, $s_b$)\\
    }
    Find $s$ best scores and corresponding sparsities of $g_c$ denoted as $\mathbb{S}_{top}$ in $\mathbb{L}$\\
    Find mean $\mu_{\mathbb{S}}$ and standard deviation $\sigma_{\mathbb{S}}$ of sparsities in $\mathbb{S}_{top}$\\
    $s_a=\max(0,\mu_{\mathbb{S}} - \sigma_{\mathbb{S}})$, $s_b=\min(1,\mu_{\mathbb{S}} + \sigma_{\mathbb{S}})$
}
Find best score and corresponding selection mask $g_{best}$ across all sparsity-refinement iterations\\
$\mathcal{F} = g_{best} \odot \mathcal{F}$\\
Return $\mathcal{F}$
\end{algorithm}

\begin{algorithm}[htbp]
\caption{FedSuccessiveHalving}
\textbf{Input}: List of feature strings $\mathcal{F}$, number of starting candidate masks $\mathcal{N}_h$, halving ratio $\gamma$, number of levels $\mathcal{L}$, number of communications in $l$-th level $\mathcal{C}_{l}$, sparsity lower bound $s_a$ and upper bound $s_b$\\
\For{each $c=1,2,...,\mathcal{N}_h$}{
    Draw random sparsity $\mathcal{S}_{c}$ from Uniform($s_a$,$s_b$)\\
    Generate random mask, $g_{c} \in \mathbb{R}^{|\mathcal{F}|}$, by drawing each element from a Bernoulli trial with probability of drawing 1 being $\mathcal{S}_{c}$.
}
$G = [g_{c} \; \forall \; c=1,...,\mathcal{N}_h$]\\
\For{level $l=1,...,\mathcal{L}$}{
    \For{each $g_c \in G$}{
        Each client constructs the numerical features based on engineered feature strings $\mathcal{F}_{c}$ on local data\\
        Train downstream model in federated fashion for for $\mathcal{C}$ server-client communication rounds with new engineered features on each client and evaluate validation score denoted as score$_{l,g_c}$
    }
    Keep top $\gamma$ of $g_c \in G$ by score$_{l,g_c}$ and remove rest from $G$\\
}
Return remaining $g_c \in G$ and their final scores $\text{score}_{\mathcal{L},g_c}$ as list of tuples $[(g_c,\text{score}_{\mathcal{L},g_c}) \; \forall \; g_c \in G]$
\end{algorithm}

We propose a federated feawture selection algorithm in Algorithm \ref{sec:fed-halving}. Our approach is inspired by Hyperband \cite{hyperband} where successive halving is used with varying resource allocations (in this case the resource is communications between client and server). First, a large set of random candidate feature masks are generated. The feature masks are vectors of 0s and 1s that determine which features to keep and which to remove. The operation of masking the feature vector is denoted as $g \odot \mathcal{F}$. For each of these feature masks, each client removes the features specified by the feature mask and then the model is trained and evaluated using FedAvg (or any other horizontal FL algorithm) but with only a very limited number of communication rounds. Then, the top performing feature masks are kept, and the rest discarded. The next round repeats the same process but with a larger number of communication rounds in the FL training. This is repeated until the best performing masks are found with a large number of communication rounds.

This process is repeated for various number of levels similarly to Hyperband. Finally, we cannot simply uniformly draw our selection masks because this results in an average sparsity of 0.5. However, we find that depending on the dataset, base AutoFE algorithm used, and degree of non-IID across clients, the best performing sparsity on the feature mask greatly differs from 0.5. Thus, we have an outer loop that gradually refines the best performing sparsity.

\begin{figure*}[htbp]
\centerline{\includegraphics[width=.99\textwidth]{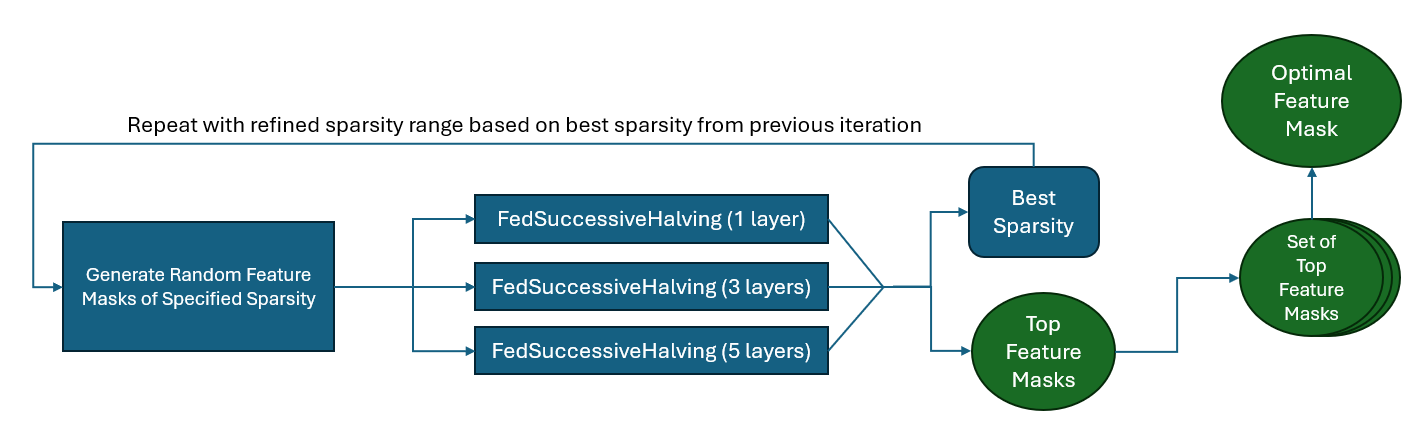}}
\caption{Workflow of Algorithm \ref{sec:fed-halving}}
\label{flafe_diagram}
\end{figure*}

We provide a simple experiment on one of the datasets studying the efficacy of this federated feature selection technique. We use OpenFE on the jungle chess dataset. OpenFE is known to create an extremely large number of features, so this is a good candidate AutoFE method to test our federated feature selection method. We perform OpenFE on each client separately then find the union of the engineered features. We perform 3 trials in the IID case (each of these trials takes a very long time as the new feature space is so large with OpenFE that evaluations for hyperparameter tuning take a very long time). The mean (std dev) base test score with raw features is 0.6804 (0.0041). For the engineered features \textit{without} federated feature selection it is 0.6349 (0.0860). For the engineered features \textit{with} federated feature selection it is 0.7654 (0.0104). This represents -6.69\% change without federated feature selection and a 12.49\% change with federated feature selection. This shows that the federated feature selection we propose in Algorithm \ref{sec:fed-halving} works quite well for AutoFE algorithms that generate many features (in the case of OpenFE with jungle chess dataset across 8 clients, there were around 850 generated features in the union). 

\end{document}